\documentclass[letterpaper, 10 pt, conference]{ieeeconf}  

\IEEEoverridecommandlockouts                              

\usepackage{graphicx} 
\usepackage[ruled,vlined,linesnumbered]{algorithm2e}
\usepackage{amsfonts, amsmath, amssymb}
\usepackage{tabularray}
\usepackage{tabularx}
\usepackage{multirow}

\usepackage{xcolor}
\usepackage{soul}

\title{\LARGE \bf
Dehazing-aided Multi-Rate Multi-Modal Pose Estimation Framework for Mitigating Visual Disturbances in Extreme Underwater Domain}

\author{Vidya Sudevan$^{1}$, 
Fakhreddine Zayer$^{1}$,
Taimur Hassan$^{2}$, 
Sajid Javed$^{1}$, \\
Hamad Karki$^{1}$, 
Giulia De Masi$^{1,3}$, 
Jorge Dias$^{1}$ 
\thanks{$^{1}$Center for Autonomous Robotic Systems, Khalifa University, Abu Dhabi, UAE.}
\thanks{$^{2}$College of Engineering, Abu Dhabi University, Al Ain, UAE.}
\thanks{$^{3}$Autonomous Robotics Research Center, Technology Innovation Institute, Abu Dhabi, UAE }
}

\begin{document}

\maketitle
\thispagestyle{empty}
\pagestyle{empty}

\begin{abstract}
This paper delves into the potential of DU-VIO, a dehazing-aided hybrid multi-rate multi-modal Visual-Inertial Odometry (VIO) estimation framework, designed to thrive in the challenging realm of extreme underwater environments. The cutting-edge DU-VIO framework is incorporating a Generative Adversarial Network (GAN)-based pre-processing module and a hybrid CNN-LSTM module for precise pose estimation, using visibility-enhanced underwater images and raw Inertial Measurement Unit (IMU) data. Accurate pose estimation is paramount for various underwater robotics and exploration applications. However, underwater visibility is often compromised by suspended particles and attenuation effects, rendering visual-inertial pose estimation a formidable challenge. DU-VIO aims to overcome these limitations by effectively removing visual disturbances from raw image data, enhancing the quality of image features used for pose estimation. We demonstrate the effectiveness of DU-VIO by calculating Root Mean Square Error (RMSE) scores for translation and rotation vectors in comparison to their reference values. These scores are then compared to those of a base model using a modified AQUALOC Dataset. Our analysis encompasses RMSE scores related to pose error, as well as an evaluation of inference speed, power consumption, GPU utilization, GPU memory usage, and temperature during the inference phase. This study's significance lies in its potential to revolutionize underwater robotics and exploration. DU-VIO offers a robust solution to the persistent challenge of underwater visibility, significantly improving the accuracy of pose estimation. We validate DU-VIO's capabilities through rigorous testing in diverse extreme underwater scenarios, showcasing its efficacy in mitigating the impact of visual disturbances on pose estimation accuracy. This research contributes valuable insights and tools for advancing underwater technology, with far-reaching implications for scientific research, environmental monitoring, and industrial applications.
\end{abstract}


\section{INTRODUCTION} \label{into}

Accurate localization of unmanned robotic platforms in GPS-denied underwater environments is crucial for autonomous operations \cite{qin2022real}. Traditional localization methods, such as dead reckoning, acoustic location, and geophysical navigation, face challenges due to electromagnetic wave attenuation underwater \cite{gonzalez2023survey}. Vision-based pose estimation suffers from errors in scenarios with low visibility, poor texture representation, and color attenuation, 
\begin{figure}[!ht]
\includegraphics[width=\columnwidth]{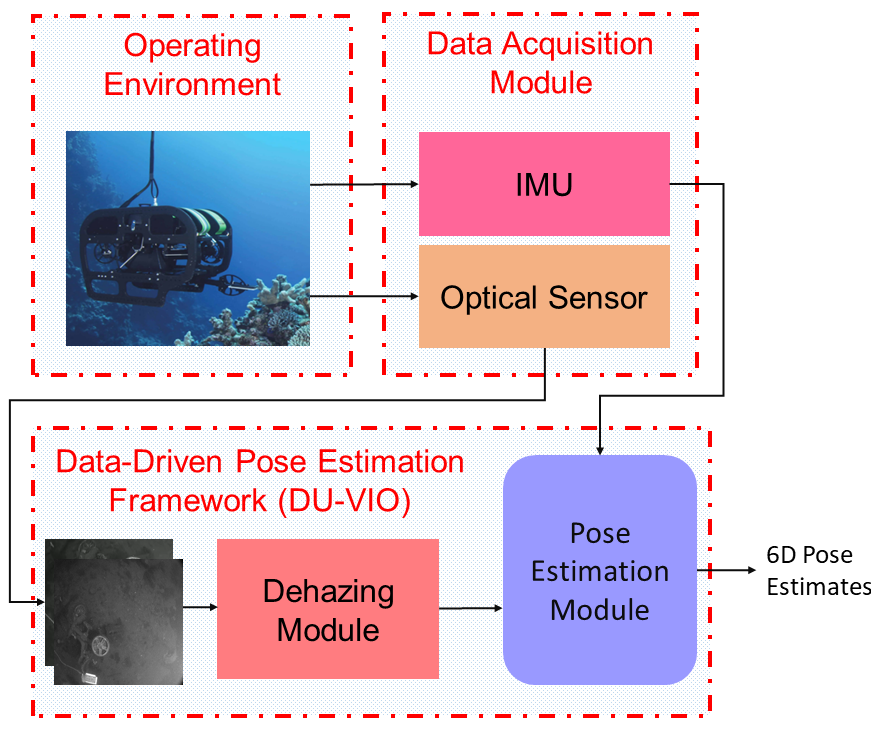}
\centering
\caption{High-level representation of DU-VIO Framework – Dehazing module reduces visual disturbances in raw camera images before feeding them, along with multi-rate IMU data, to the pose estimation module for translation and orientation estimation.}
\label{fig:high_level}
\end{figure}
common in extreme underwater conditions \cite{quattrini2017experimental}. Inertial Measurement Units (IMUs) used for underwater navigation experience drift and pose prediction inaccuracies over time \cite{jalal2021underwater}. Combining monocular cameras and IMUs enhances pose estimation accuracy \cite{clark2017vinet}.
\begin{figure*}[!phbt]
\includegraphics[width=\textwidth]{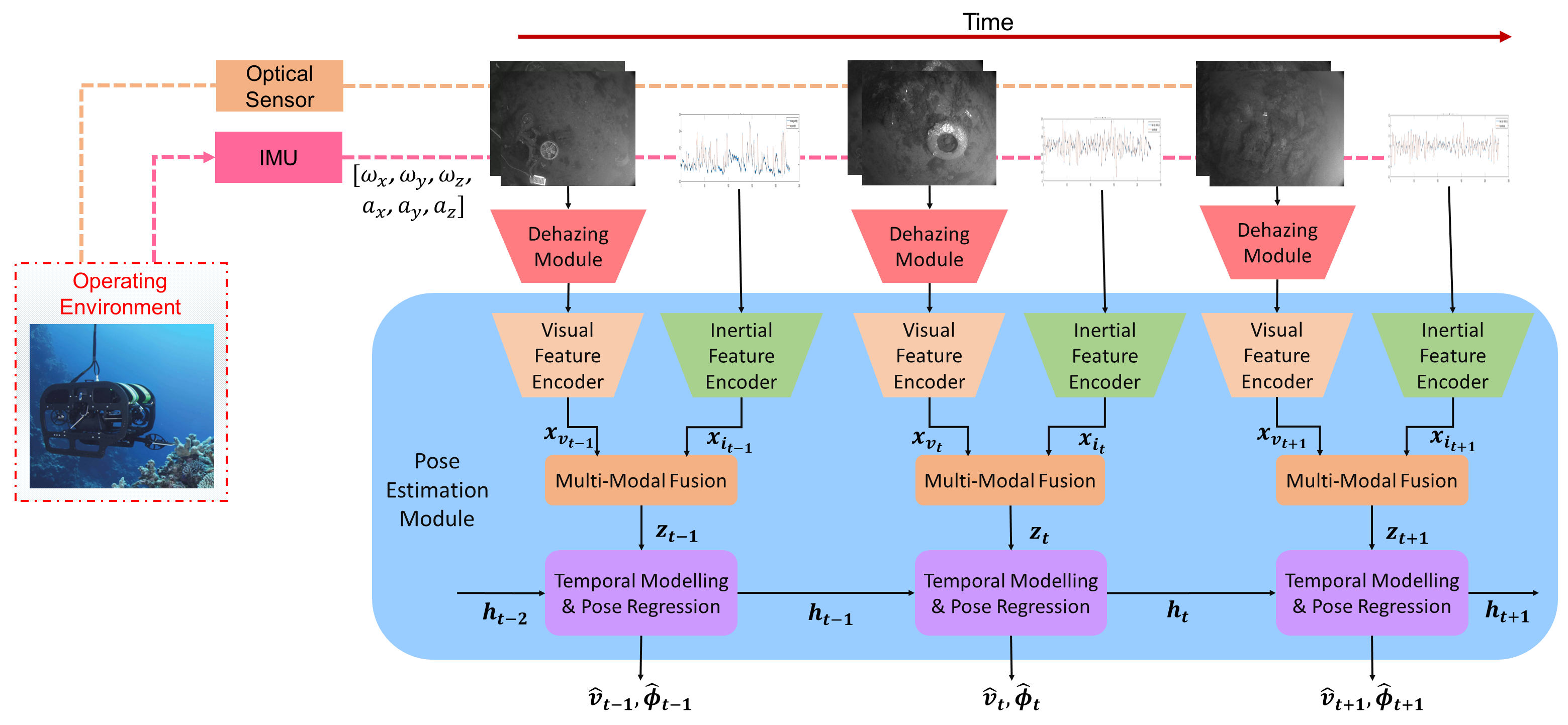}
    \centering
    \caption{DU-VIO Framework Overview: Raw camera images are dehazed to improve visibility. Visual features are extracted with a visual feature encoder, while inertial features are extracted from unprocessed IMU data. The two sets of features are fused using a multimodal fusion module, and the 6D pose is estimated with a temporal modeling and pose regression module.}
    \label{fig:overall}
\end{figure*}
Visual-Inertial Odometry (VIO) estimates vehicle pose using camera and IMU data. Geometry-based VIO methods use feature detection, matching, motion estimation, outlier rejection, scale estimation, and pose optimization \cite{sudevan2022multisensor} \cite{zhao2021ego}, but their real-world adaptability is limited \cite{qin2022real}. Data-driven VIO frameworks employ CNNs and RNNs to estimate camera egomotion, optical flow parameters, and extract high-level features from images and IMU data \cite{almalioglu2022selfvio}. However, tailored learning-based multi-rate multi-modal posture estimation frameworks for underwater environments remain unexplored.

The proposed Dehazing-aided Underwater Visual-Inertial Odometry (DU-VIO) Framework was conceptualized in response to the facts mentioned above and considering the visual disturbances caused by the extreme levels of turbidity, distorted and low-textured images and the lack of dedicated learning-based multi-rate multi-modal-pose estimation frameworks in the underwater domain. The high-level representation of the DU-VIO framework is depicted in Figure \ref{fig:high_level}. Development of the DU-VIO framework is based on the assumption that the integration of a dehazing module with the VIO framework will improve the accuracy of pose estimation and its adaptability to mitigate the visual-disturbance challenges in the extreme underwater environment and, thereby, improving the accuracy of pose estimation.\\
This research introduces DU-VIO, a novel data-driven VIO framework with a dehazing module, evaluated in challenging underwater conditions with visual disturbances like distortion, turbidity, and low-textured images. We build upon the VS-VIO framework \cite{yang2022efficient}, known for its performance on the KITTI dataset. To adapt to low-textured underwater environments, we replace the policy network with a visibility enhancement module. DU-VIO preprocesses raw images using a GAN-based dehazing module and estimates six-dimensional pose with a hybrid CNN and LSTM architecture. Testing involved three scenarios using the modified AQUALOC dataset, comparing DU-VIO with and without the dehazing module. U-VIO refers to DU-VIO without dehazing. Figure \ref{fig:overall} depicts the DU-VIO framework.\\
In summary, our approach makes the following major contributions:
\begin{itemize}
    \item Pioneers in proposing a dedicated learning-based dehazing-aided multi-rate multi-modal hybrid CNN-LSTM framework for accurate pose estimation in challenging underwater environments. These environments are characterized by distortions, turbidity, and low-textured image captures.
  \item Evaluation of the effectiveness of data-driven DU-VIO frameworks for underwater pose estimation, considering three distinct underwater visual disturbances, leveraging a modified version of the AQUALOC Dataset.
\item Utilization of the Root Mean Square Error (RMSE) metric to quantitatively assess translation and rotation errors between predicted pose estimates and ground truth across various scenarios.
\item Comprehensive hardware evaluation metrics for the DU-VIO framework, including inference speed, GPU power consumption, GPU utilization, GPU memory usage, and temperature, are meticulously documented in this research.
\end{itemize}
This paper is organized as follows: Section \ref{sota} provides an overview of state-of-the-art techniques in underwater image enhancement and visual-inertial pose estimation. Section \ref{du-vio} outlines the development of the DU-VIO framework, emphasizing its dehazing-aided capabilities for precise visual-inertial pose estimation. Section \ref{network_training} delves into the training specifics of the DU-VIO framework. Section \ref{results_analysis} presents comprehensive experimental results. Finally, in Section \ref{conclusion}, we draw conclusions based on our findings and contributions.

\section{State-of-the-Art Approaches} \label{sota}
Due to the degraded image quality caused by suspended particles and medium properties, enhancing image visibility is essential for underwater visual tasks. In highly turbid underwater environments, light absorption increases, diminishing the visual perception capability. In addition, backscattered light causes severe distortions, making it impossible to distinguish between features. This section concentrates on data-driven visibility enhancement techniques and visual-inertial odometry (VIO) techniques for underwater applications, as the research aim is to develop a dehazing-aided learning-based VIO framework.

\subsection{Visibility Enhancement in Harsh Environments}

The Simultaneous Localization and Mapping (SLAM) application and visibility improvement methods have recently been combined in the underwater domain. In \cite{cho2017visibility} and \cite{huang2019underwater}, the Contrast-Limited Adaptive Histogram Equalization (CLAHE) method and Retinex theory-based color correction were combined with SLAM to improve underwater image quality and enhance its performance. It has been reported that these methods achieved only a marginal improvement and were ineffective under extremely turbid conditions \cite{zheng2023real}.  

On the other hand, the utilization of Generative Adversarial Network (GAN) based Image-to-Image (I2I) translation algorithms presents the potential to improve textile and content representations, resulting in the generation of realistic images that exhibit distinct and reliable features, particularly when operating under extremely turbid environments \cite{islam2020underwater}. In \cite{chen2019real} and \cite{zheng2023real}, a combination of CycleGAN with ORB-SLAM and GAN with ORB-SLAM2 architecture is presented to take advantage of the superior performance of GAN-based approaches. The existing literature does not include reports on integrating a complete learning-based Visual-Inertial Odometry (VIO) framework with a visibility enhancement module to improve pose estimation performance in highly challenging underwater conditions. Motivated by this research direction, a GAN-based visibility enhancement framework is utilized in the proposed DU-VIO framework to pre-process the raw input images acquired from an extreme underwater environment before passing to the hybrid CNN-LSTM module for pose estimation.

\subsection{Data-Driven Visual-Inertial Odometry Approaches}

Visual-Inertial Odometry (VIO) uses images and inertial data to compensate for the errors due to rapid motion and address the sub-optimal image capture. The VIO techniques can primarily be divided into geometric-based and data-driven frameworks. Geometric VIO frameworks \cite{mourikis2007multi, ding2023rd, leutenegger2015keyframe, qin2018vinsmono} make use of handcrafted characteristics to determine the geometric relationship between extracted visual and inertial features. It involves many parallel processing blocks, parameter configurations, and complex computations. Also, geometric-based techniques are unreliable in dynamic illumination, featureless surroundings, and indistinct images. With so many manually adjustable parameters, it is impossible to model the actual world accurately. In this case, a VIO algorithm's configurations that function well in one context could provide unfavorable results in another \cite{wang2017deepvo}. Learning-based VIO algorithms can extract and provide reliable feature representations using a well-represented dataset, even in challenging conditions \cite{aslan2022hvionet}.

VINet \cite{clark2017vinet}  introduced the first data-driven VIO approach by utilizing LSTM networks for IMU data and FlowNet \cite{dosovitskiy2015flownet} for optical flow features. DeepVIO \cite{han2019deepvio} employs LSTM-based IMU  pre-integrated and fusion networks and estimates the pose by minimizing self-motion constraint loss. SelfVIO \cite{almalioglu2022selfvio} estimates egomotion and depth maps using adversarial training, adaptive vision-inertial sensor fusion and GAN enhancements. Visual-Selective-VIO (VS-VIO) \cite{yang2022efficient} selectively deactivates visual modality based on motion and IMU data. The Gumbel-Softmax approach ensures differentiability and end-to-end training. VS-VIO framework is the first fully developed open-source hybrid CNN-LSTM algorithm for learning-based VIO. The literature demonstrates that while the VIO problem has been solved in open-air settings, a learning-based multi-modal pose estimation framework for extreme underwater environments has not yet been developed. These findings emphasize the necessity for a robust learning-based system to anticipate underwater vehicle position and orientation relative to their operational environment, especially in challenging underwater environments.

\section{DEHAZING AIDED UNDERWATER VISUAL-INERTIAL ODOMETRY (DU-VIO) FRAMEWORK} \label{du-vio}

\begin{figure*}[!phbt]
\includegraphics[width=\textwidth]{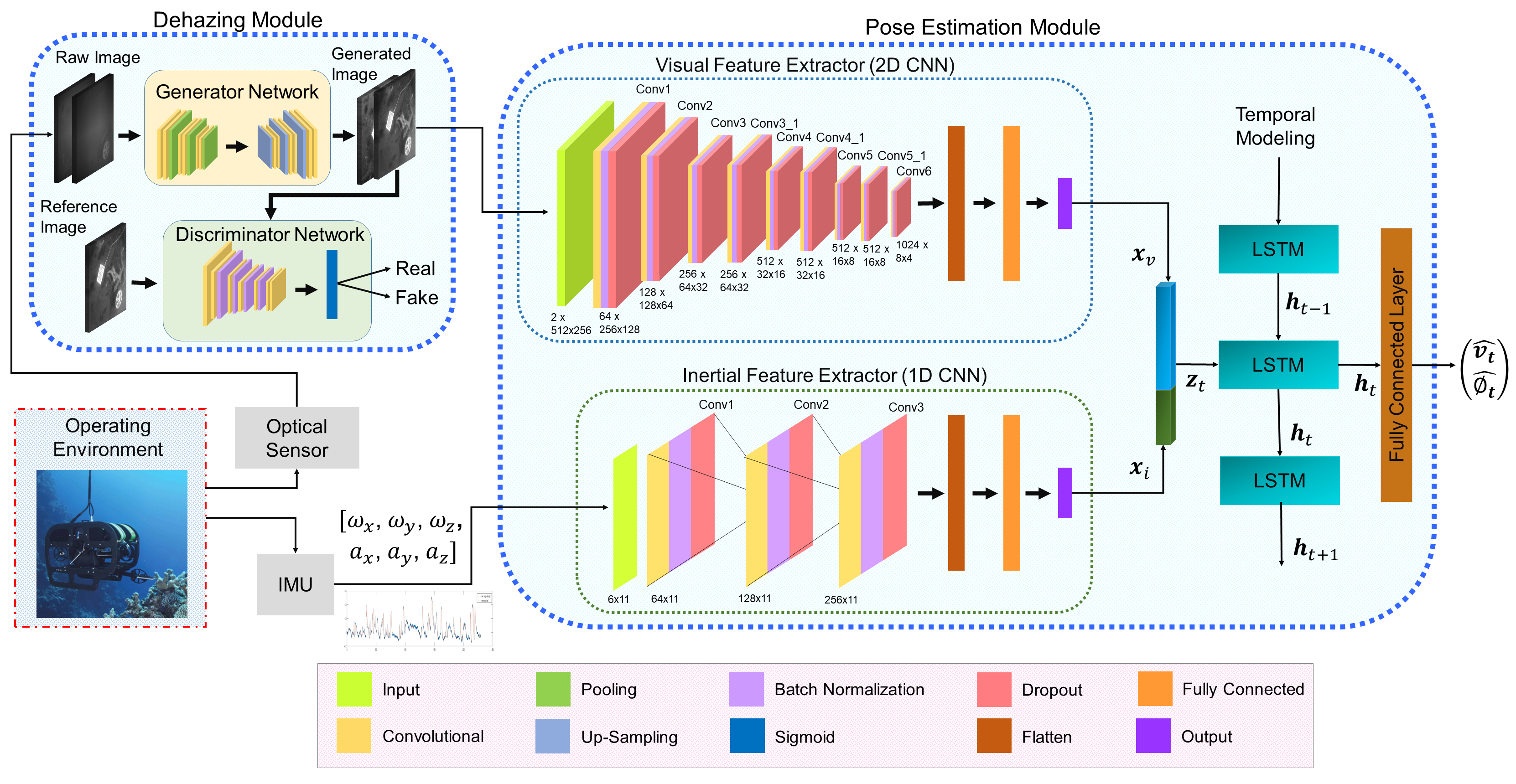}
    \centering
    \caption{Detailed Illustration of the Dehazing-aided Underwater Visual-Inertial Odometry (DU-VIO) Framework}
    \label{fig:DU_VIO block}
\end{figure*}

The Dehazing-aided Underwater Visual-Inertial Odometry (DU-VIO) modifies the VS-VIO, a learning-based multi-modal pose estimate network, without a policy network and by adding a dehazing module to estimate the pose in extreme underwater scenarios, characterized by distortion, turbidity and low-textured image captures. The VS-VIO framework adaptively selects the visual modality for pose estimation using the policy network. The suspended particles cause poor lighting and significant turbidity in the dynamic underwater environment. This makes it difficult to employ in a VIO framework intended for a demanding environment with fewer functionality due to environmental constraints. As the images captured from the real underwater environment are visually degraded, merely using the a learning-based framework alone cannot address the challenging scenarios. These unavoidable factors has to be considered while developing a VIO framework suitable for real underwater environment, which leads to the development of DU-VIO by excluding the policy network and including the visibility enhancement module. Figure \ref{fig:DU_VIO block} presents the schematic illustration of the proposed end-to-end learning-based multi-modal DU-VIO framework for pose estimation in extreme underwater domain that are invariant to visual disturbances. The details of each module is depicted below.

\subsection{Dehazing Module}
The dehazing module uses a GAN-based architecture to mitigate the visual disturbances associated with the raw images captured from the operating environment. A pre-trained Densenet-121 model, modified as an encoder-decoder structure with skip connections, is used for feature extraction in the generator network. The encoder part extracts the multi-scaled visual features from the raw images, which are then reconstructed using the decoder part by gradually increasing the spatial dimensions utilizing a series of transposed convolutional layers. The discriminator model, which outputs the probability of the generated image, includes multiple convolution layers followed by the convolution layer, batch normalization, and LeakyReLU activation functions \cite{ahmed2023vision}. These visibility enhanced images are fed to the visual feature encoder for visual feature extraction.

\subsection{Visual Feature Encoder}
The visual feature encoder module uses a two-dimensional CNN model and a fully connected layer at the network's end to extract visual features from two successive pre-processed image frames. Since the location in the current frame is closely related to the previous frame, two successive image frames are combined and fed to the CNN network for efficient learning and to estimate the pose accurately. Since the significant geometric features must be learned using the visual feature encoder \((\mathit{E_{visual}})\), the FlowNetSimple \cite{dosovitskiy2015flownet} model has been used as the feature encoder. FlowNetSimple is a nine-layered, two-dimensional CNN model used to extract features suitable for optical flow predictions. With stride two for the first six convolutional layers, the receptive fields gradually shrink from \(7\texttt{x}7\) to \(5\texttt{x}5\) and \(3\texttt{x}3\), respectively. A Leaky Rectified Linear Unit (ReLU) nonlinearity follows each convolutional layer. The features from the final convolution layer are linearized and fed to a fully connected layer to extract the required visual features, \(\mathit{\boldsymbol{x_{v}}}\).
\begin{equation}
     \mathit{\boldsymbol{x_{v}}}=\mathit{E_{visual}}(\mathbf{V_{n}},\mathbf{V_{n+1}})
    \label{equation:image_features}
\end{equation}

\subsection{Inertial Feature Encoder}
Inertial data streams possess a distinct temporal component and are typically characterized by higher frequencies (around 100 Hz) compared to images (around 10 Hz). The inertial feature encoder, \((\mathit{E_{inertial}})\), consists of three one-dimensional convolutional layers and a fully connected layer, which collectively extracts the inertial features vector. A Leaky ReLU nonlinearity follows each convolutional layer. The features obtained from the last convolutional layer are transformed into a linear format and then passed through a fully connected layer to extract the necessary inertial features. The vector \(\mathit{M_{imu}}\) represents the collection of IMU measurements, i.e., the linear acceleration and angular velocity, that have been recorded during the time interval between two consecutive image frames, \(\mathbf{V_{n}}\) and \(\mathbf{V_{n+1}}\). The measurements are inputted directly into the inertial feature encoder to extract the inertial features \(\mathit{\boldsymbol{x_{i}}}\).
\begin{equation}
     \mathit{\boldsymbol{x_{i}}}=\mathit{E_{inertial}}(\mathit{M_{imu}})
    \label{equation:inertial_features}
\end{equation}

\subsection{Multi-Modal Fusion Module}
The visual and inertial feature extractors are used to extract high-level features, which are then combined through a fusion function called \(\mathit{g_{concat}}\). In this study, the visual features \((\mathit{\boldsymbol{x_{v}}})\) and inertial features \((\mathit{\boldsymbol{x_{i}}})\) are concatenated to generate a unified feature vector \(\mathit{\boldsymbol{z_{t}}}\). This combined feature vector is subsequently utilized as input for the temporal modeling and pose regression module.  
\begin{equation}
    \mathit{\boldsymbol{z_{t}}}=\mathit{g_{concat}}( \mathit{\boldsymbol{x_{v}}}, \mathit{\boldsymbol{x_{i}}})
    \label{equation:concat_features}
\end{equation}

\subsection{Temporal Modelling and Pose Regression Module}
An RNN network is used to learn the inter-dependencies present in the sequence of motions and to capture the sequential nature of the concatenated feature vector \(\mathit{\boldsymbol{z_{t}}}\). The RNN network used for temporal modeling and pose regression contains a two-layered LSTM network,  followed by a two-layered Multi-Layer Perceptron (MLP) network. The 6-DoF pose estimation is performed by passing the hidden state of the final LSTM layer through the MLP network at each time step. The subsequent sections provide comprehensive information on the training process of the U-VIO architecture, the dataset used, and the experimental outcomes.
\begin{equation}
     \left (\mathit{\boldsymbol{h_{t}}}, \widehat{\mathit{\boldsymbol{v_{t} }}}, \widehat{\mathit{\boldsymbol{\phi_{t} }}} \right )= RNN\left ( \mathit{\boldsymbol{z_{t}}}, \mathit{\boldsymbol{h_{t-1}}} \right )
    \label{equation:pose}
\end{equation}
where \(\mathit{\boldsymbol{h_{t}}}\) and \(\mathit{\boldsymbol{h_{t-1}}}\) represents the hidden latent vectors of RNN at time \(\mathit{t}\) and \(\mathit{t-1}\) respectively.

\section{TRAINING OF POSE ESTIMATION FRAMEWORKS} \label{network_training}
This section describes DU-VIO framework training under extreme underwater conditions. The AQUALOC dataset \cite{ferrera2019aqualoc} includes monocular pictures, IMU data, pressure sensor data, and offline structure-from-motion library ground truth pose estimates from three natural underwater environments. The DU-VIO framework is trained and evaluated using seven sequences from 'harbor site' subset of AQUALOC dataset, \(\left \{ h01,h02,\cdot \cdot ,h07 \right \}\).

\subsection{Underwater Multi-Modal Dataset}
As detailed in \cite{ferrera2019aqualoc}, the ground truth trajectory poses are computed offline using the Colmap library from a subset of images (1 out of 5 images was used). It has to be noted that the sequences h02, h04, and h07 contain 39, 55, and 6 missing ground truth instances, respectively. With these facts, each of the seven ground truth trajectory sequence poses is linearly interpolated and used for the training and inference phase. 

\begin{figure}[!ht]
\includegraphics[width=\columnwidth]{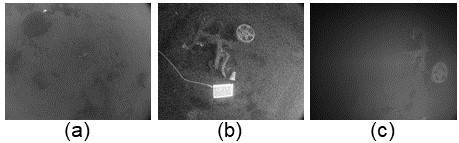}
\centering
\caption{Scenarios: (a) Original (b) Distortion, and (c) Turbid}
\label{fig:scenario}
\end{figure}

As the research objective is to evaluate the effectiveness of the DU-VIO framework to estimate the pose in extreme underwater scenarios by mitigating the visual disturbances, three distinct visual disturbances are considered. The low-texture scenario is regarded as the original scenario due to the fact that sequences of images from the AQUALOC dataset already contain images that characterize the low-texture scenario. The publicly available noise models are utilized to add additional distortion and turbidity effects to the images. For the ease of computation, only one-third of the data from each trajectory sequences were considered for pose estimation. Sample images from each scenario are presented in Figure \ref{fig:scenario}.

\subsection{Parameters of DU-VIO Framework} 
The modified AQUALOC dataset is used to train the proposed DU-VIO framework, with and without dehazing module. AQUALOC images are 20 Hz, but IMU data are 200 Hz. Images, IMU data, and reference poses are not synchronized. The parameter selection for the dehazing module is similar to \cite{ahmed2023vision}. For the pose estimation module, the monochromatic images are scaled to \(512\texttt{x}256\) pixels and, two consecutive image frames are fed to the visual feature extractor, to extract high-level visual features of size \(512\) from \(2\texttt{x}512\texttt{x}256\) visual input. 11 IMU readings occurring between two consecutive image frames are fed to the inertial feature encoder to extract the high-level inertial feature vector of size \(256\texttt{x}11\) from \(6\texttt{x}11\) inertial input vector.The visual and inertial feature vectors are concatenated and fed to a two-layer LSTM with \(1024\) hidden units per layer for pose estimation. A two-layered MLP network exploits the hidden state of the final LSTM layer to estimate the 6-DoF pose at each time step. A batch size of \(16\) and the learning rate of \(1\times 10^{-6}\) is selected. The batch size is set to \(16\) and the learning rate chosen for training the DU-VIO framework is \(1\times 10^{-6}\). The Adam optimizer with \(\alpha = 0.9\) and \(\beta =0.999\) is used due to due to its comparatively lower memory demand \cite{kingma2014adam}. The Mean Squared Error (MSE) loss function is utilized during the training process in order to minimize errors in translational and rotational pose estimation. 

\begin{equation}
    \mathit{L_{pose}}=\frac{1}{T-1}\sum_{t=1}^{T-1}\left ( \left \| \widehat{\mathit{\boldsymbol{v_{t}}}}-\mathit{\boldsymbol{v_{t}}} \right \|_{2}^{2} + \alpha \left \| \widehat{\mathit{\boldsymbol{\phi_{t}}}}-\mathit{\boldsymbol{\phi_{t}}} \right \|_{2}^{2} \right )
    \label{equation:mse}
\end{equation}
where \(T\) is the sequence length, \(\mathit{\boldsymbol{v_{t}}}\) and \(\mathit{\boldsymbol{\phi_{t}}}\) represents the reference values of translation and rotational vectors. As explained in \cite{yang2022efficient}, the parameter \(\alpha \) represents the weight to balance the translational and rotational loss. 

\subsection{Details of Computing Resources}
The dehazing module developed using TensorFlow 2.1.0 library and is implemented on Core i9-10940@3.30 GHz processor, single NVIDIA Quadro RTX 6000 GPU. To effectively remove the visual disturbances, 80\% whole images from the modified AQUALOC dataset was used to train the dehazing module foe 50 epochs. The pose estimation module was developed using PyTorch 1.12.1 library and is implemented on Google Colab Pro Plus with the A100 GPU. As the trajectory sequences are provided in the increasing order of complexity, the trajectory sequences \(\left \{ h02,h04,h06 \right \}\) are used for training, \(\left \{ h03,h05 \right \}\) are used for validation and the sequences \(\left \{ h01,h07 \right \}\) are used for testing the DU-VIO framework under original, distortion and turbid scenario for 20 epochs. For the DU-VIO framework without the dehazing module, it took 2.5 hours to train each scenario, whereas it took 4 hours to train the DU-VIO framework with the dehazing module.  

\section{EXPERIMENTAL RESULTS AND DISCUSSIONS} \label{results_analysis}
Using the sequences h01 and h07 from the 'harbour site' modified subset of the AQUALOC dataset, the efficacy of the DU-VIO framework in mitigating visual disturbances in extreme underwater environments is evaluated. These sequences were selected due to the complex nature of the dataset's representation, which includes abrupt motion changes and the temporary absence of visual information caused by collisions. Compared to sequence h01, sequence h07 is more overexposed and abrupt \cite{qin2022survey}. The Root Mean Square Error (RMSE) metric is utilized to compare the performance DU-VIO under various scenarios. 

\begin{table} [!ht]
\caption{Generator Model: Ablation Study}
\label{table:ablation}
\centering
\begin{tabular}{|l|l|l|l|l|} 
\hline
Backbone              & \textbf{PSNR}    & \textbf{SSIM}   & \textbf{MSE}    & \textbf{RMSE}    \\ 
\hline
\textbf{DenseNet-121} \cite{huang2017densely} & \textbf{26.8726} & \textbf{0.9612} & \textbf{158.47} & \textbf{12.588}  \\ 
\hline
\textbf{ResNet-50} \cite{jian2016deep}    & 26.1324          & 0.9308          & 203.29          & 14.258           \\ 
\hline
\textbf{ViT} \cite{dosovitskiy2020image}          & 26.4283          & 0.9475          & 176.31          & 13.278           \\ 
\hline
\textbf{MobileNet-v2} \cite{sandler2018mobilenetv2} & 25.5721          & 0.9026          & 242.75          & 15.580           \\ 
\hline
\textbf{VGG-16} \cite{simonyan2014very}      & 25.3408          & 0.8859          & 263.18          & 16.222           \\
\hline
\end{tabular}
\end{table}

Table \ref{table:ablation} presents the ablation experimental results to analyze the capability of the dehazing module with different backbone models. The DenseNet-121 model is chosen to implement the dehazing module due to its superior performance across all evaluation matrices considered here. The performance evaluation of the dehazing module with the state-of-the-art algorithm is depicted in Table \ref{table:dehaze_benchmark}. 

\begin{table} [!ht]
\caption{State-of-the-art comparison of dehazing module}
\label{table:dehaze_benchmark}
\centering
\begin{tabular}{|l|l|l|l|l|} 
\hline
Framework               & \textbf{PSNR}    & \textbf{SSIM}   & \textbf{MSE}    & \textbf{RMSE}    \\ 
\hline
\textbf{Proposed}       & \textbf{26.8726} & \textbf{0.9612} & \textbf{158.47} & \textbf{12.588}  \\ 
\hline
\textbf{FDA} \cite{yang2020fda}             & 24.0709          & 0.8807          & 249.23          & 15.787           \\ 
\hline
\textbf{DehazeFormer-B} \cite{song2023vision}  & 25.5284          & 0.9357          & 196.81          & 14.028           \\ 
\hline
\textbf{FFA-Net} \cite{qin2020ffa}       & 24.9351          & 0.9063          & 224.62          & 14.987           \\
\hline
\end{tabular}
\end{table}

\begin{figure}[!ht]
\includegraphics[width=\columnwidth]{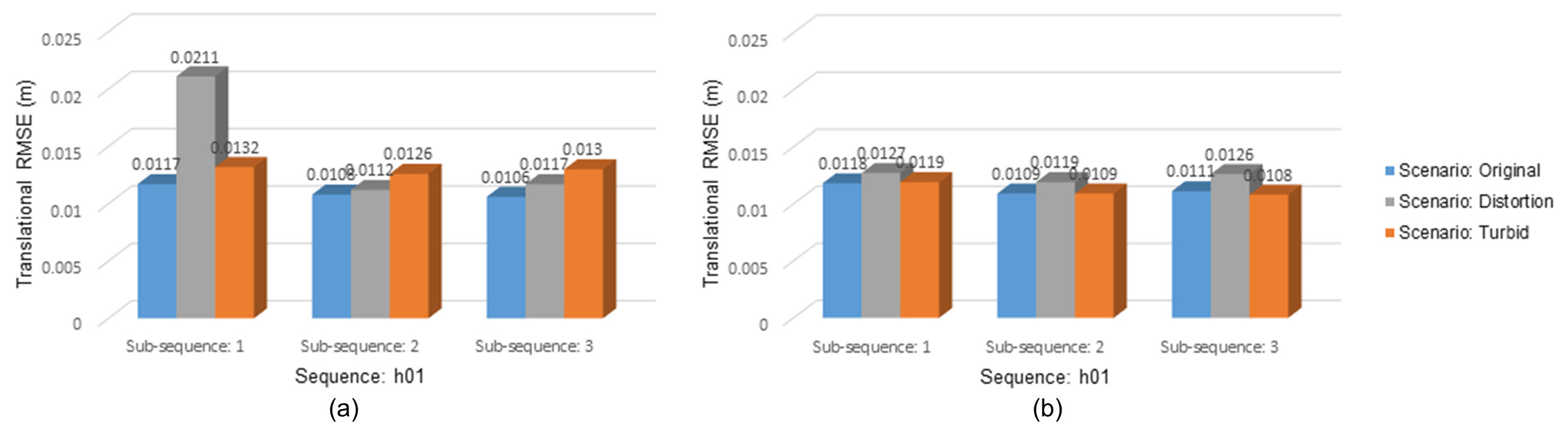}
\centering
\caption{Translation RMSE (\(\boldsymbol{v _{rmse}}\)) scores for sequence: h01 (a) Without dehazing module, and (b) With dehazing module}
\label{fig:h1_trans}
\end{figure}

\begin{figure}[!ht]
\includegraphics[width=\columnwidth]{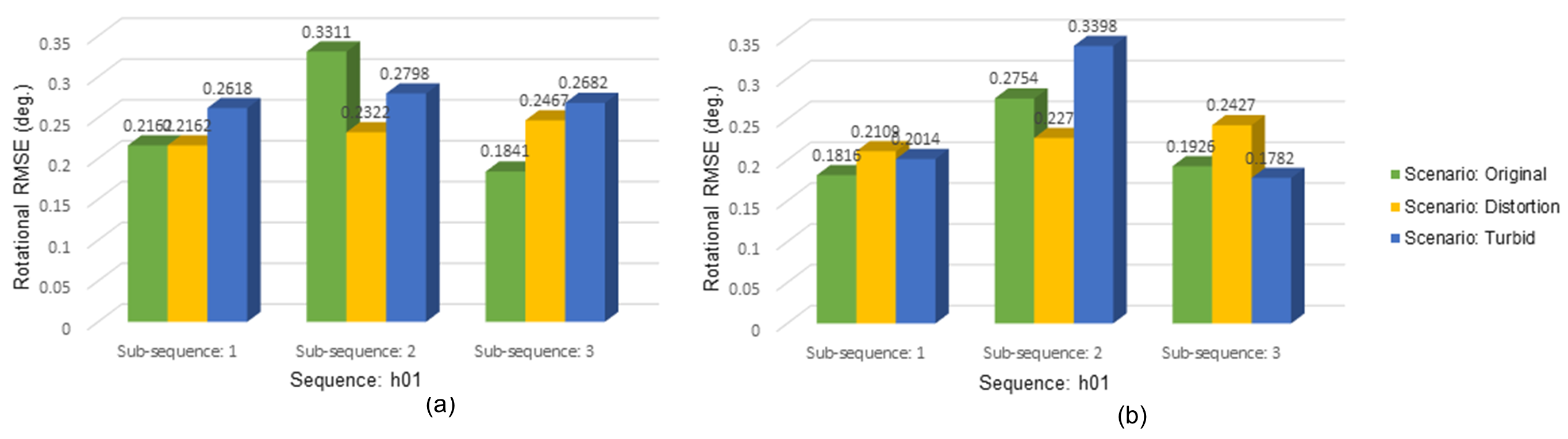}
\centering
\caption{Rotational RMSE(\(\boldsymbol{\phi _{rmse}}\)) scores for sequence: h01 (a) Without dehazing module, and (b) With dehazing module}
\label{fig:h1_rotate}
\end{figure}

\begin{figure}[!ht]
\includegraphics[width=\columnwidth]{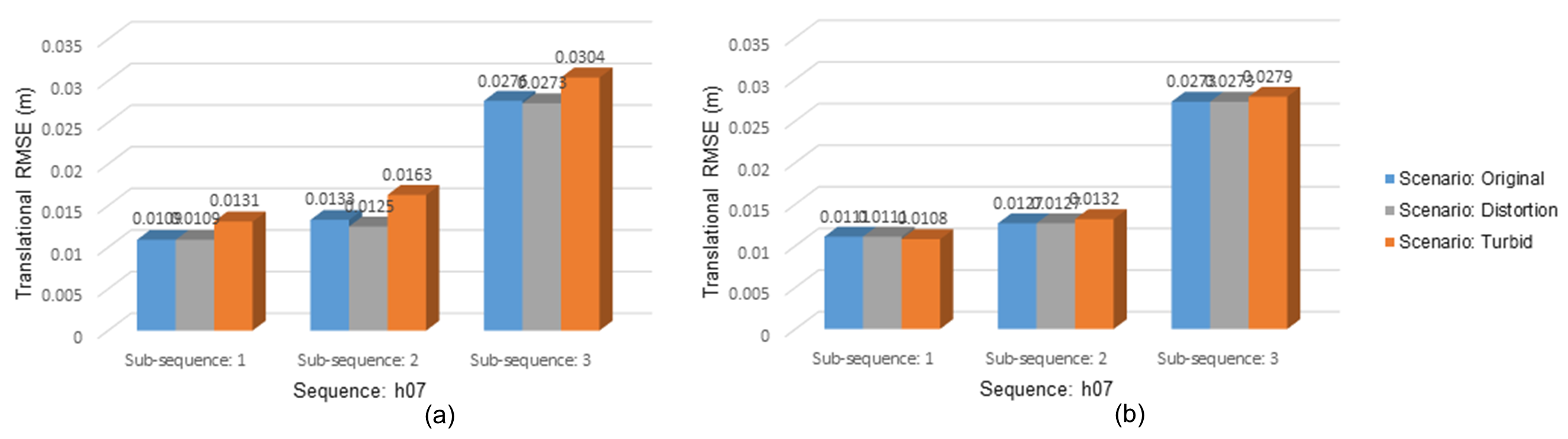}
\centering
\caption{Translation RMSE (\(\boldsymbol{v _{rmse}}\)) scores for sequence: h07 (a) Without dehazing module, and (b) With dehazing module}
\label{fig:h7_trans}
\end{figure}

\begin{figure}[!ht]
\includegraphics[width=\columnwidth]{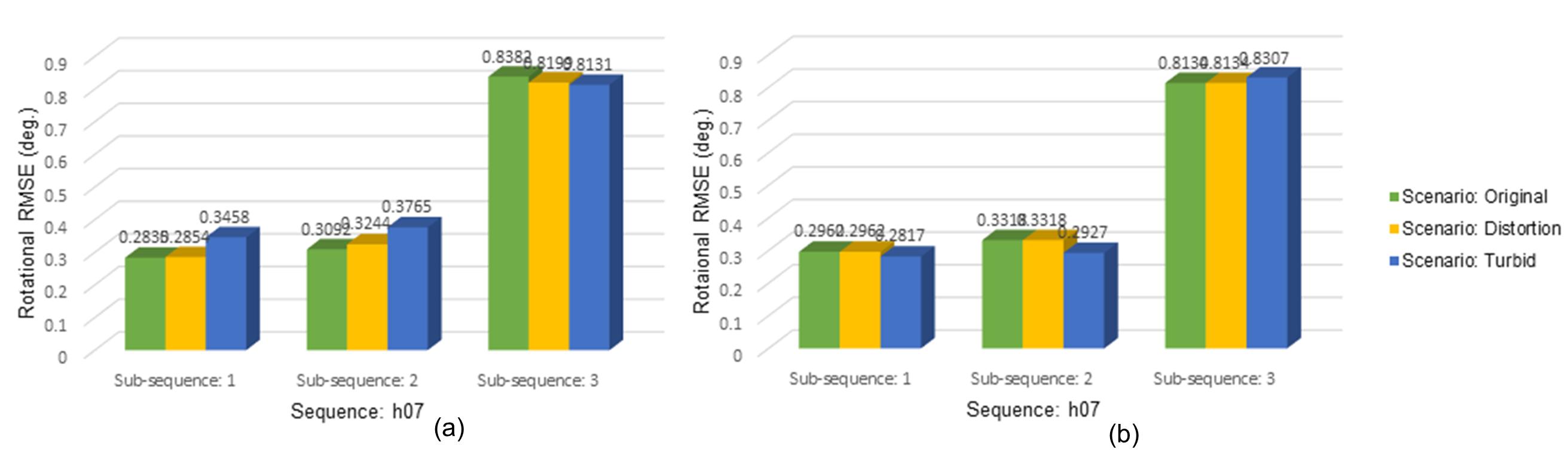}
\centering
\caption{Rotational RMSE(\(\boldsymbol{\phi _{rmse}}\)) scores for sequence: h07 (a) Without dehazing module, and (b) With dehazing module}
\label{fig:h7_rotate}
\end{figure}

For pose estimation, the selected h01 and h07 sequences are equally divided into three sub-sequences, and calculated the translational and the rotational RMSE scores with respect to the reference pose vector under all three scenarios. The RMSE scores obtained for the translational and rotational vectors of sequence h01, without and with the use of dehazing module is illustrated in Figure \ref{fig:h1_trans} and Figure \ref{fig:h1_rotate}, respectively. In most of the cases, with the use of dehazing module, the RMSE scores are minimized and are matching or even lesser than the original scenarios. Figure \ref{fig:h7_trans} and Figure \ref{fig:h7_rotate} represent the RMSE scores associated with the translational and rotational vectors, respectively, for sequence h07 without and with the use of dehazing module. 

\begin{table}[!ht]
\caption{State-of-the-art comparison of DU-VIO Framework}
\label{table:benchmark}
\centering
\begin{tblr}{
  row{1} = {c},
  cell{1}{1} = {c=2}{},
  cell{2}{1} = {r=2}{},
  cell{2}{3} = {c},
  cell{2}{4} = {c},
  cell{3}{3} = {c},
  cell{3}{4} = {c},
  cell{4}{1} = {r=2}{},
  cell{4}{3} = {c},
  cell{4}{4} = {c},
  cell{5}{3} = {c},
  cell{5}{4} = {c},
  vlines,
  hline{1-2,4,6} = {-}{},
  hline{3,5} = {2-4}{},
}
\textbf{Frameworks } &                    & \textbf{Sequence: h01} & \textbf{Sequence: h07} \\
{Geometry\\Based}    & \textbf{OKVIS} \cite{leutenegger2015keyframe}     & 0.0406                 & 0.1171                 \\
                     & \textbf{ORB-SLAM3} \cite{campos2021orb} & 0.0198                 & 0.0212                 \\
{Data\\Driven}       & \textbf{VINet} \cite{clark2017vinet}     & 0.0497                 & 0.1495                 \\
                     & \textbf{DU-VIO} [ours]   & \textbf{0.0111}        & \textbf{0.0188}        
\end{tblr}
\end{table}

\begin{table}[!ht]
\caption{Hardware Metrics}
\label{table:hardware}
\centering
\begin{tabular}{|l|r|} 
\hline
\textbf{Inference Time (s)}                     & 40     \\ 
\hline
\textbf{Power Consumption (W)~ ~ ~~}            & 47.41  \\ 
\hline
\textbf{GPU Usage (\%)}                         & 4\%    \\ 
\hline
\textbf{Memory Used (MiB)}                      & 923    \\ 
\hline
\textbf{GPU Temperature (\textsuperscript{o}C)} & 34     \\
\hline
\end{tabular}
\end{table}

Most of the experimental results shows the ability of DU-VIO framework to estimate the pose in underwater environment irrespective of the visual disturbances. The DU-VIO framework demonstrates its ability to learn the temporal relationship between multi-modal visual-inertial data representations and accurately predict the pose estimates in challenging underwater environments. This capability is evident in both the simple h01 sequence and the challenging h07 sequence. This could be due to the interpolation of the data used to train the network, as the reference pose presented in the original dataset utilizes only a subset of the images for computation. The interpolation allowed the visual-inertial encoder to derive high-level features from two consecutive frames. The hardware evaluation metric for the DU-VIO framework is presented in Table \ref{table:hardware}. The algorithmic performance of DU-VIO framework with dehazing module under original scenario is compared with the state-of-the-art VIO frameworks and is presented in Table \ref{table:benchmark}.

\section{CONCLUSIONS} \label{conclusion}
This paper introduced DU-VIO, an innovative dehazing-assisted hybrid multi-rate multi-modal Visual-Inertial Odometry (VIO) framework tailored to overcome the complexities of extreme underwater environments with various visual disturbances. DU-VIO's integration of a GAN-based preprocessing module to mitigate visual disturbances and a hybrid CNN-LSTM module for enhanced pose estimation, utilizing both dehazed images and raw IMU data, demonstrated outstanding performance in challenging underwater scenarios. Experimental validation using the modified AQUALOC dataset across multiple scenarios reaffirmed the framework's robustness. Additionally, this work provides valuable insights for potential enhancements in DU-VIO, contributing to the continuous advancement of pose estimation algorithms for extreme underwater environments. The research represents a significant stride towards enabling more precise and reliable underwater exploration and navigation.

\section*{ACKNOWLEDGMENT}

The authors acknowledge the project “Heterogeneous Swarm of Underwater Autonomous Vehicles”, a collaborative research project between the Technology Innovation Institute (Abu Dhabi)  and Khalifa University  (contract no. TII/ARRC/2047/2020). This work is also supported by Khalifa University under Awards No. RC1-2018-KUCARS-8474000136. 



\end{document}